\documentclass[letterpaper, 10 pt, conference]{ieeeconf}  

\IEEEoverridecommandlockouts                              

\usepackage{graphicx}
\usepackage{subfigure}
\usepackage{amsmath}
\usepackage{booktabs}
\usepackage{multirow}
\usepackage{multicol}
\usepackage{cite}
\usepackage[colorlinks,
            linkcolor=blue,
            anchorcolor=blue,
            citecolor=blue,
            urlcolor=blue]{hyperref}
\usepackage{geometry}

\title{\LARGE \bf
PS6D: Point Cloud Based Symmetry-Aware 6D Object Pose Estimation in Robot Bin-Picking
}

\author{Yifan Yang$^{1}$, Zhihao Cui$^{2}$, Qianyi Zhang$^{1}$ and Jingtai Liu$^{1*}$
\thanks{This work was done during internship of Yifan Yang at Mech-Mind. This work is supported by Mech-Mind and National Natural Science Foundation of China under Grant 62173189.}
\thanks{$^{1}$Yifan Yang, Qianyi Zhang and Jingtai Liu are with the Institute of Robotics and Automatic Information System, Nankai University, Tianjin, China.  
        {\tt\small Emails: 2012773@mail.nankai.edu.cn; 1120210190@mail.nankai.edu.cn; liujt@nankai.edu.cn.}}%
\thanks{$^{2}$Zhihao Cui is with the Deep Learning Group, Mech-Mind, Beijing,
China. 
        {\tt\small Email: cui.zhihao@mech-mind.net}}%
\thanks{*Corresponding author}
}

\begin{document}
\newgeometry{left=55pt,right=55pt,top=55pt,bottom=58pt}

\maketitle
\thispagestyle{empty}
\pagestyle{empty}

\begin{abstract}

6D object pose estimation holds essential roles in various fields, particularly in the grasping of industrial workpieces. Given challenges like rust, high reflectivity, and absent textures, this paper introduces a point cloud based pose estimation framework (PS6D). PS6D centers on slender and multi-symmetric objects. It extracts multi-scale features through an attention-guided feature extraction module, designs a symmetry-aware rotation loss and a center distance sensitive translation loss to regress the pose of each point to the centroid of the instance, and then uses a two-stage clustering method to complete instance segmentation and pose estimation. Objects from the Siléane and IPA datasets and typical workpieces from industrial practice are used to generate data and evaluate the algorithm. In comparison to the state-of-the-art approach, PS6D demonstrates an 11.5\% improvement in F$_{1_{inst}}$ and a 14.8\% improvement in Recall. The main part of PS6D has been deployed to the software of Mech-Mind, and achieves a 91.7\% success rate in bin-picking experiments, marking its application in industrial pose estimation tasks.

\end{abstract}

\section{Introduction}

The use of industrial robotic arms for object grasping plays a crucial role in various manufacturing and automation applications. The ability to automate the picking and placing of objects is fundamental to achieving efficiency, speed, and precision in industrial processes. In the field of 6D pose estimation, traditional methods \cite{r1,r2} often exhibit poor performance when dealing with occluded and stacked objects. Currently, the academic community is increasingly focusing on deep learning based 6D pose estimation methods. However, most of these methods predominantly rely on RGB \cite{r6,r7,r8,r9,r10,r11} or RGB-D \cite{r12,r13,r14,r15,r16,r17,r18,r19} information because they require the rich features provided by RGB information for subsequent predictions. While these methods might be effective in scenarios involving daily object grasping, they struggle with industrial object grasping becauseof the following reasons. Firstly, surface rust and color variations in different batches often introduce interference to RGB information. Also, highly reflective surfaces, low contrast, and the absence of texture make it challenging for networks based on RGB information to extract useful features. For some black objects, manually annotating RGB datasets can be extremely difficult. 

\begin{figure}[tb]	
	\centering
	\subfigure[]{
        \label{changtiao1}
		\includegraphics[height=0.31\linewidth]{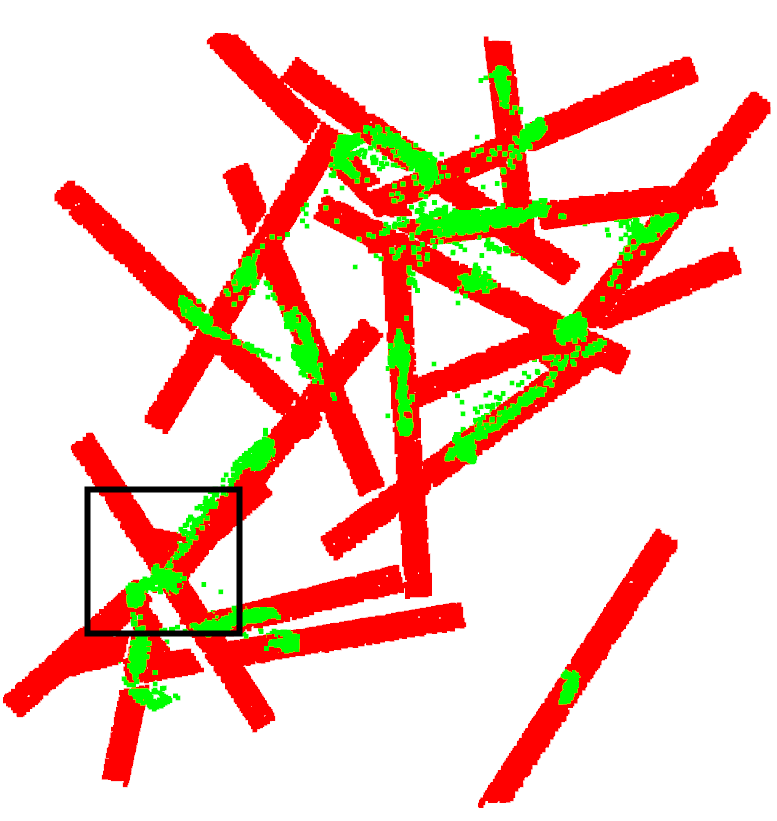}}
	\subfigure[]{
        \label{changtiao2}
		\includegraphics[height=0.31\linewidth]{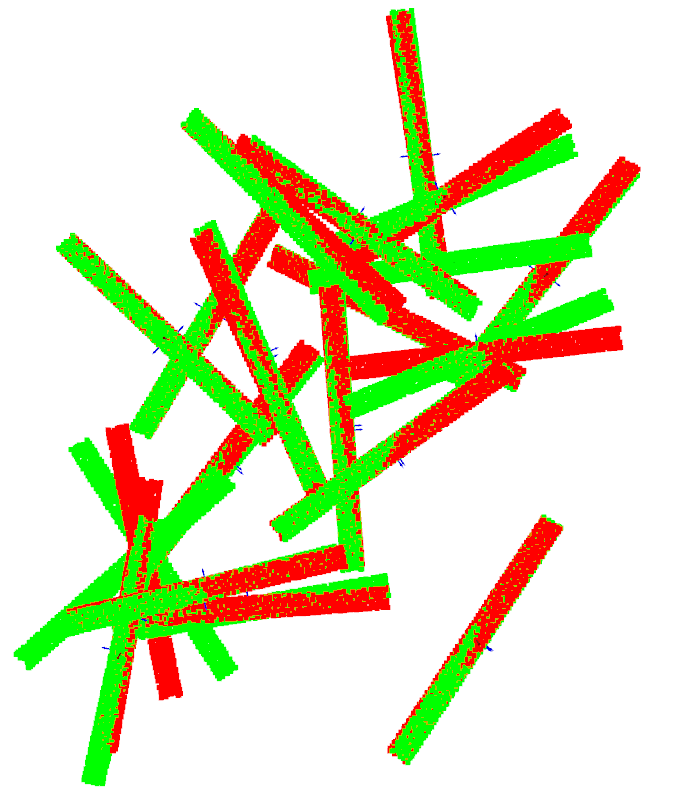}}
	\subfigure[]{
        \label{24_01}
		\includegraphics[height=0.31\linewidth]{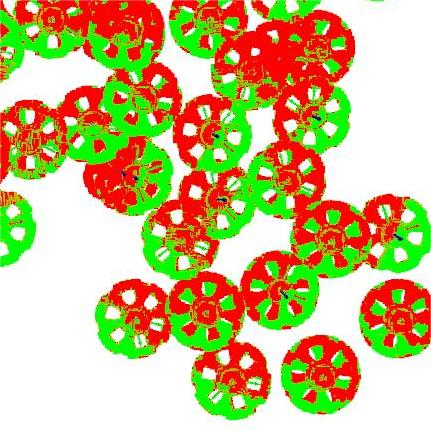}}
	\caption{Existing algorithms encounter problems. (a) and (b) show the occurrence of missed detections when slender objects intersect, and the inability to accurately estimate pose; (c) shows the problem of large rotational regression bias for multi-symmetric objects. In (a), the red point cloud represents the real poses, while the green points represent the predicted centroids. In (b) and (c), the red point cloud represents the predicted poses, and the green point cloud represents the real poses.}
\end{figure}
There is limited targeted research on symmetrical objects (especially infinitely symmetrical objects) and slender objects, which happen to be common in industrial settings. In existing algorithms, such as PPR-Net \cite{r25}, single-stage clustering based on predicted centroid position information, is commonly used for pose voting. However, this approach faces two problems. One is that in cases where the intersection point is close to the center of the slender object as illustrated in Fig. \ref{changtiao1}, clustering algorithms may struggle to differentiate them, leading to missed detection. Since the predicted results of two objects are clustered together, the averaged result is likely to be inconsistent with either object, which is shown in Fig. \ref{changtiao2}. The other problem is that, for objects with finite symmetry, due to the existence of multiple symmetric rotation matrices, points on the same instance tend to have different rotation predictions and may be clustered into the same category. This leads to an averaged pose that is different from any of the equivalent poses, resulting in incorrect rotation predictions as illustrated in Fig. \ref{24_01}. 

It should also be realized that most of the objects in existing public datasets are taken from daily life, but not closely related to industrial practical applications, which also makes it necessary to construct a robot bin-picking dataset based on real workpieces from industrial sites.


In response to the above challenges, we propose a point cloud based symmetry-aware instance-level 6D pose estimation method (PS6D). First, we normalize the scene point cloud and place it within a 100mm$\times$100mm $\times$100mm space to mitigate the impact of variations in object sizes and camera parameters on model performance. Next, we construct an attention-guided feature extraction module, which obtains multi-scale global and local feature. Extracted features are then fed into the center distance sensitive centroid regression module and the symmetry-aware rotation prediction module. The normalized translation and rotation information from the network output are concatenated and entered two-stage clustering, resulting in an optimal pose for each object. Overall, PS6D outperforms the state-of-art approach in simulation. The main part of PS6D has been deployed in the software of Mech-Mind for the task of estimating the pose of workpieces in actual industrial scenarios.

In summary, the main contributions of this work are as follows.

\begin{enumerate}
    \item A novel point cloud based symmetry-aware 6D object pose estimation framework (PS6D) is introduced, which outperforms state-of-art approaches.
    \item A center distance-sensitive translation loss and a symmetry-aware rotation loss are designed in PS6D, and a two-stage clustering method is proposed to enhance the accuracy of pose estimation, particularly for slender and multi-symmetric objects.
    \item The PS6D dataset is constructed and evaluated, incorporating five typical industrial objects to better align with real-world industrial applications.
\end{enumerate}
\section{Related Work}
\subsection{RGB Methods}
RGB methods use color information captured by a single RGB camera for object pose estimation. These algorithms typically rely on in-depth analysis and feature extraction of color information. They train deep neural networks to recognize and match feature points in the images, enabling the estimation of the object's pose. Due to the use of only RGB information, these algorithms are sensitive to changes in lighting conditions and color variations. However, they offer advantages in terms of computational efficiency and real-time performance. For example, SSD-6D \cite{r6} extends the popular SSD paradigm to cover the entire 6D pose space, training exclusively on synthetic model data. PoseCNN \cite{r7} estimates the 3D translation of an object by locating its center in the image, predicting its distance from the camera, representing rotation with quaternions, and introducing a loss function capable of handling symmetric objects. \cite{r8} directly predicts the 2D image positions of the projection vertices of the object's 3D bounding box. \cite{r9} introduces a pixel-level voting network (PVNet) that regresses pixel-level vectors pointing to key points and uses these vectors to vote for key point locations. \cite{r10} solves the pose ambiguity problem without predefined symmetries, inspired by the VSD (Visible Surface Difference) metric. \cite{r11} uses an encoder-decoder network to predict correspondences between densely sampled pixels and fragments, estimating the 6D pose of rigid objects with available 3D models from a single RGB input image.

\subsection{RGB-D Methods}
Algorithms based on RGB-D information utilize images captured by RGB-D cameras to acquire both color and depth information of objects for pose estimation. These algorithms combine color and geometric information, performing joint analysis on RGB images and depth maps using deep neural networks to achieve more accurate and robust pose estimation. PointFusion \cite{r12} processes image data and raw point cloud data separately using Convolutional Neural Network (CNN) and PointNet \cite{pointnet}. DenseFusion \cite{r13} proposes a method to combine color and depth information in RGB-D inputs, and integrates the iterative optimization process into the neural network architecture. PVN3D \cite{r14} develops a deep Hough voting network for detecting 3D key points of objects, and estimates 6D pose parameters using least squares fitting. \cite{r15} introduces a bidirectional fusion module, taking into account object texture and geometry information, and designs a simple yet effective 3D key point selection algorithm. \cite{r16} designs a symmetric invariant pose distance metric, utilizing XYZNet to efficiently extract pointwise features from RGB-D data and directly regressing 6D pose. \cite{r17} introduces rotation and object bounding box confidence-aware prediction, and proposes a new point-wise object bounding box voting mechanism. \cite{r18} improves category-level object pose and size prediction by modifying pose consistency learning. \cite{r19} uses a deep multi-directional fusion network to efficiently fuse RGB-D frames from multiple perspectives.

\subsection{Point Cloud Methods}
In the industrial workpiece grasping scenario, due to the existence of the situation in Section \uppercase\expandafter{\romannumeral1}, RGB information may cause interference, so it is necessary to study some methods that only use point clouds to complete pose estimation. \cite{r20} uses self-attention mechanisms to bring geometric features together, and employ a deep Hough voting scheme to generate pose suggestions. \cite{r21} estimates 6D object pose from a single depth image and target object mask based on a model-based approach. \cite{r22} designs a novel single-shot approach based entirely on synthetic data. \cite{r23} treats parameterized shapes as a category, estimate pose by least squares fitting of individual instance and template key points with centroids. \cite{r24} jointly predicts feasible 6-DoF grasping poses, instance semantic segmentation, and collision information, achieving an end-to-end differentiable framework. PPR-Net \cite{r25} and PPR-Net++ \cite{r26} are most similar to our work, as they can perform point cloud instance segmentation and pose estimation simultaneously through centroid regression and Euler angle rotation prediction. However, they perform poorly when dealing with slender and multi-symmetric objects.
\section{Method}
\begin{figure*}[tb]
  \centering
  \includegraphics[height=0.49\linewidth]{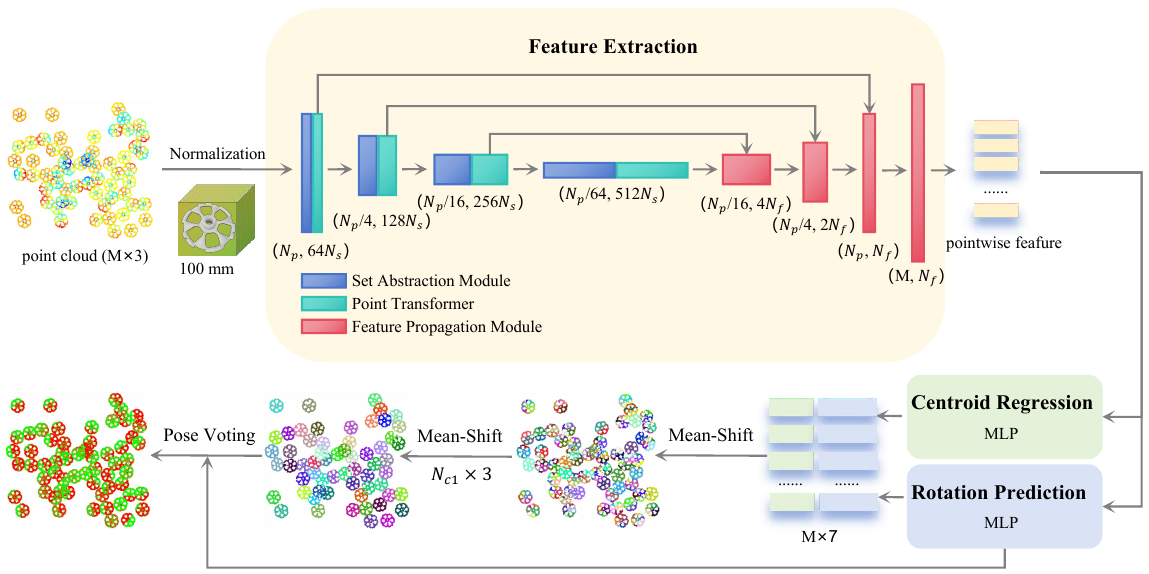}
  \caption{Architecture of PS6D. 
The network takes only the per-point position information as input. It performs feature extraction on the normalized point cloud, predicts translation and rotation, and finally undergoes two clustering stages and pose voting. This process ultimately achieves instance segmentation and pose estimation.}
  \label{network}
\end{figure*}
\subsection{PS6D Network Architecture}
Our network architecture is illustrated in Fig. \ref{network}. It takes normalized point clouds as input. Through an attention-guided feature extraction module, the network obtains multi-scale global and local features. These features are then fed into the centroid regression module and rotation prediction module, enabling point-wise predictions of quaternions and offsets relative to the centroid. The network's loss function can be expressed as equation \ref{loss}, where $L_t$ and $L_r$ are translation loss and rotation loss respectively, with $W_t$ and $W_r$ representing the corresponding loss weights.
\begin{equation}
    L=W_r L_r+W_t L_t
    \label{loss}
\end{equation}
\noindent\textbf{Attention-Guided Feature Extraction Module \quad}PointNet++ \cite{1} has been proven to be a leading method of point cloud feature extraction, and is widely used as a backbone in other state-of-the-art methods. In this paper, we introduce the Point Transformer structure \cite{2} based on PointNet++. Due to the inherent permutation invariance of the self-attention mechanism, it can handle unordered point cloud data without being affected by the order of points, which makes it reasonable to apply Point Transformer to process point cloud data.

In our PS6D network, we employ the Set Abstraction module to downsample point clouds of size $M \times 3$. Subsequently, we connect Point Transformer modules layer by layer, facilitating the network to dynamically learn relationship weights between different points. This adaptability enables the model to handle point cloud data with diverse shapes, densities, and distributions, progressively extracting both global and local features. Lastly, we utilize the Feature Propagation module for upsampling, completing the feature fusion and transfer process, ultimately producing features with a dimension of $N_f$.

\noindent\textbf{Symmetry-Aware Rotation Prediction Module \quad}In 6D object pose estimation, symmetry is a significant concern as it leads to a challenging problem of multiple solutions due to the object producing similar projections under different poses. The camera observations may correspond to multiple poses of the object, introducing ambiguity that negatively affects training.

\begin{figure}[tb]	
	\centering
	\subfigure[IPARingScrew]{
        \label{IPARingScrew}
		\includegraphics[height=0.31\linewidth]{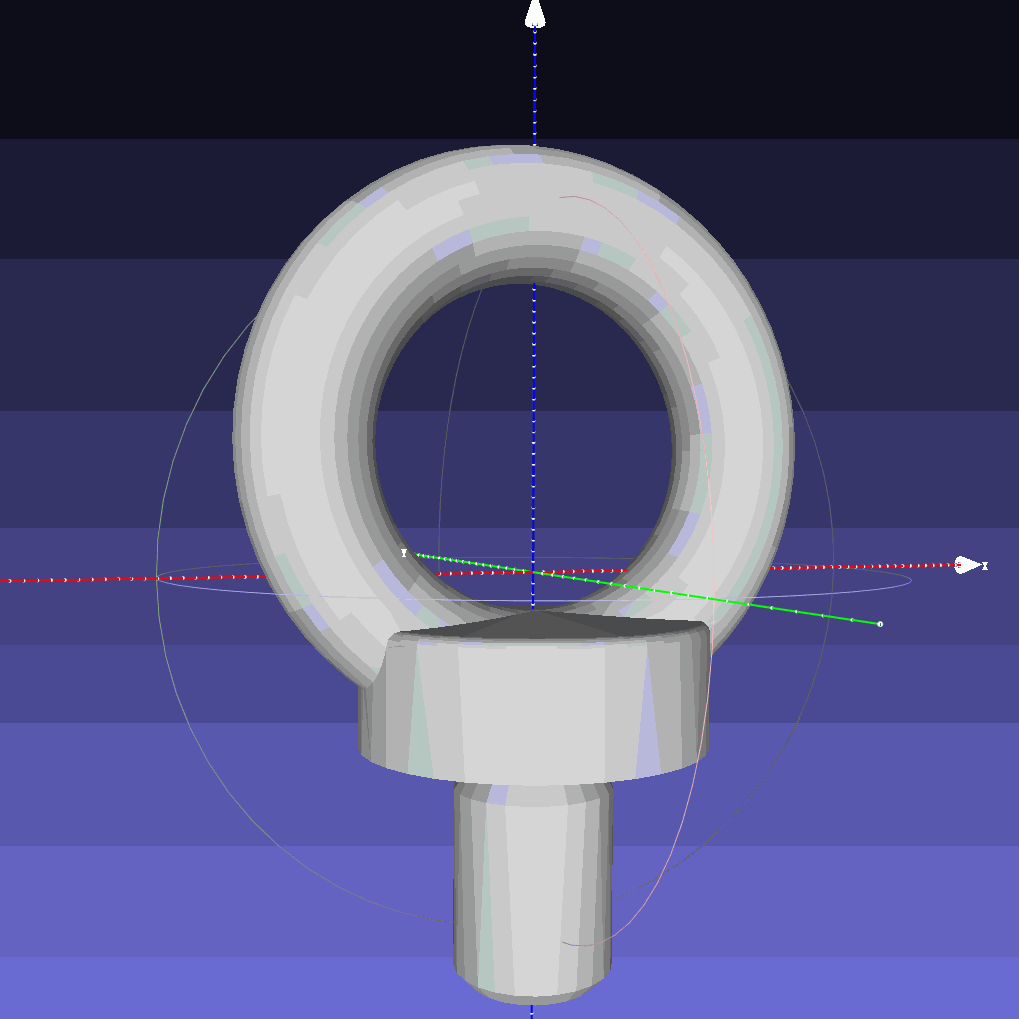}}
	\subfigure[Bunny]{
        \label{Bunny}
		\includegraphics[height=0.31\linewidth]{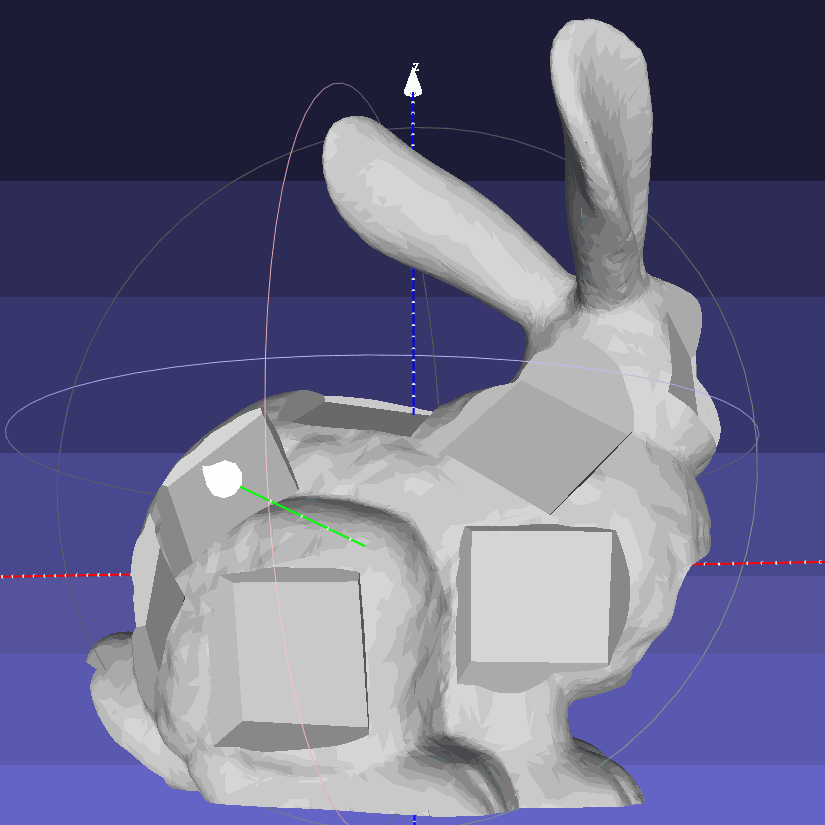}}
    \subfigure[Candlestick]{
        \label{Candlestick}
		\includegraphics[height=0.31\linewidth]{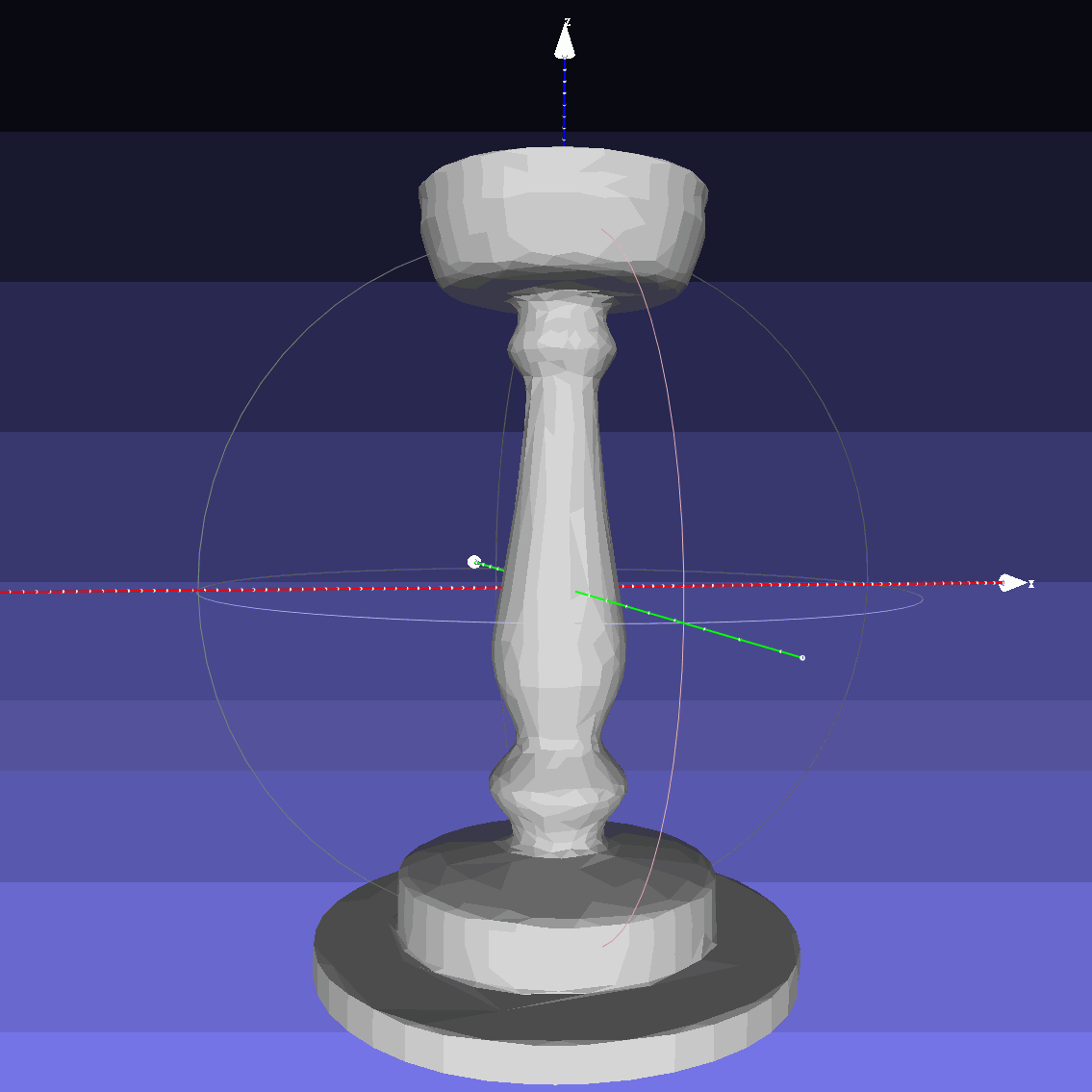}}
	\caption{Three models with different symmetries. (a) with d$_x$=d$_y$=0, d$_z$=180, and S=[
[  [1, 0, 0],
  [0, 1, 0],
  [0, 0, 1]
],
[  [-1, 0, 0],
  [0, -1, 0],
  [0, 0, 1]
]], (b) without symmetry, (c) with infinite symmetry around the z-axis and no symmetry around the other axes.}
	\label{object_examples} 
\end{figure}

To address this issue, we propose a symmetry-aware rotation loss function. Before training, it is essential to record the symmetry information of the object, denoted as $d_x$, $d_y$, $d_z$, and $T_s$. Here, $d_x$, $d_y$, and $d_z$ represent the degrees of symmetry around the xyz axes, as illustrated in Fig. \ref{IPARingScrew}. $T_s$ represents the threshold for infinite symmetry, indicating that the object is considered infinitely symmetric when the symmetry degrees exceed $T_s$. In our implementation, we set $T_s$ to 15 degrees. For parts with finite symmetry, we calculate a list of symmetric rotation matrices $S$. Specifically, for objects with no symmetry (e.g., Fig. \ref{Bunny}) or objects with only infinite symmetry (e.g., Fig. \ref{Candlestick}), this list contains only the identity matrix. In representing object rotations, compared to Euler angles, quaternions do not suffer from singularities or gimbal lock, and offer higher computational efficiency in subsequent mathematical operations. Therefore, we choose to use quaternions for regression. Due to the non-connectivity of the quaternion space, when designing rotation loss, we first convert quaternions to rotation matrices for calculation. 

Objects with infinite symmetry are abstracted as directed line segments located at the infinite symmetric axis (for a spherical object, it is abstracted as the center point). We select the rotation matrix $s$ from $S$, which minimizes the distance between the predicted model point cloud and the ground truth model point cloud. This optimal symmetry matrix is then used to compute the rotation loss in equation \ref{rot_loss}, where $M$ represents the point cloud set of objects in the scene, $n$ is the number of objects in the scene. $R^{gt}$ and $R^{pred}$ are the real and predicted rotation matrices respectively, and $v$ represents the infinite symmetry of the object. Except for the dimension corresponding to infinite symmetry, which is set to 1, the other two dimensions in vector $v$ are set to 0. Specifically, if the object does not possess infinite symmetry, then $v$ is a vector of all ones. If the object is a sphere, $v$ is zero vector.
\begin{equation}
    L_r=\frac{1}{n}\sum_{M_i \in M} \min_{s \in S} \Vert R_i^{gt} \times s \times ( M_i \cdot v ) - R_i^{pred} \times ( M_i \cdot v ) \Vert
    \label{rot_loss}
\end{equation}

\noindent\textbf{Center Distance Sensitive Centroid Regression Module \quad}This module aims to regress the point clouds of objects in the scene to their respective instance centroids. For certain slender objects, such as PS6D\_23 with a bounding box of [32 1304 92] mm, regressing points located at the ends, far from the centroid, can be more challenging. Based on this, we designed a center distance sensitive translation loss function (as shown in equation \ref{trans_loss}), where $M$ is the point cloud set of objects in the scene, $n$ is the number of objects in the scene, $T^{gt}$ and $T^{pred}$ represent the real and predicted instance centroid positions, and $C_i$ is a matrix obtained by normalizing the distance of each point to its centroid within the [0.5, 1.5] range for the i-th instance.
\begin{equation}
    L_t=\frac{1}{n}\sum_{i \in [1, |M|]} \Vert ( T_i^{gt} - T_i^{pred} ) C_i \Vert
    \label{trans_loss}
\end{equation}
\subsection{Normalized Workpiece Space}
In practical industrial applications of object grasping, different objects often vary significantly in size. Due to constraints on the radius of the SA module and other network parameters, a single set of network parameters may struggle to adapt to objects of different sizes. Therefore, it is necessary to normalize the point clouds of workpieces. The specific steps are as follows. First, normalize all object model point clouds to a cubic space with a side length of 100mm$\times$100mm $\times$100mm. This ensures that a single set of network parameters can be applied to objects of various sizes. Next, move the scene point cloud to the coordinate origin to mitigate the impact of camera position on model performance.
\subsection{Two-Stage Clustering}
To solve the problem of inaccurate pose prediction for slender and multi-symmetric objects, we introduce a two-stage clustering method. The input for the first stage clustering consists of the predicted centroid position information (xyz, $M \times 3$) and per-point rotation information (quaternions, $M \times 4$) from the network. It considers both translation and rotation, effectively distinguishing between intersecting objects. For symmetric objects, the number of categories $N_{c1}$ generated by the first stage clustering is generally greater than the number of instances $n$ in the scene. Ideally, $N_{c1}=n\times len(S)$. The input for the second stage clustering is the average position of the cluster centroids generated in the previous stage ($N_{c1} \times 3$), and the parameters are typically set such that the bandwidth is smaller than the previous stage and the minimum number of points is greater than the previous stage. Since only position information is considered at this stage, it is possible to merge the poses belonging to the same instance generated in the previous stage into one category. The quaternion with the smallest error within the category is selected as the predicted rotation in the pose voting step, resulting in the final output of $N_{c2}\times 7$ predicted poses.
\section{Experiment}
In this section, we aim to test our model using datasets suitable for robot industrial grasping scenes, design appropriate evaluation metrics, and compare our model with existing methods. 

\subsection{Implementation Details}
PS6D is deployed in Pytorch. Each training epoch consists of 30 scenes, and each scene contains 60 point clouds. The training data size for each point cloud is $32768\times 3$. The initial learning rate is set to 0.001, and after 12 epochs, the learning rate is decayed by a factor of 0.7 at each step, reaching a minimum of 1e-6. We set $N_p=4096$, $N_f=128$, and $W_r=W_t=1$. Both training and testing were conducted using an NVIDIA GeForce RTX 3060 GPU. The forward propagation time for one scene's point cloud is 16ms.

\subsection{Dataset Selection and Generation}
Currently, there are many datasets related to 6D pose estimation, but most of them are not specifically designed for industrial grasping scenarios. Objects in datasets like LineMOD \cite{3} and YCB-Video \cite{r7} are commonly seen in daily life, and they often have rich texture features. However, industrial grasping objects are typically silver-white workpieces, often with high reflectivity and a lack of texture information. T-Less dataset \cite{4} has a small amount of data, leading to strong correlations between samples and making it difficult to simulate scenarios of scattered and stacked objects in industrial grasping.

Taking all these considerations into account, we choose to use partial objects from the Siléane dataset \cite{5} and IPA dataset \cite{6} to test our method. To make the data more representative of industrial scenarios, we select five typical workpieces used in industrial settings to construct the PS6D dataset and carry out relevant metric tests. 

Given the CAD models of the objects, we use Blender to simulate the process of objects randomly falling into a bin from a certain height, preserving physical collisions to achieve the effect of objects scattered and stacked in the bin. In previous work of Mech-Mind, a method for generating a dataset for 6D pose estimation in industrial grasping has already been developed, and our work is based on it. As generating visibility information for each instance during data generation is time-consuming, and our network does not include a visibility calculation module, we use sparsely stacked data (either directly generating such data or sparsifying heavily stacked data using data augmentation) during training.

\begin{figure}[tb]	
	\centering
	\subfigure[PS6D\_6]{
		\includegraphics[width=0.4\linewidth]{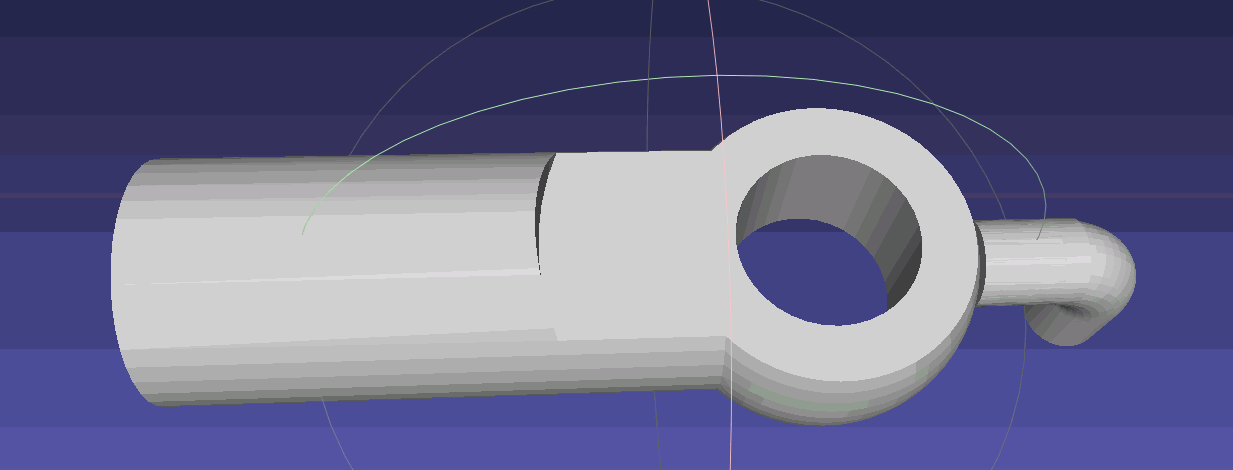}}
	\subfigure[PS6D\_23]{
		\includegraphics[width=0.4\linewidth]{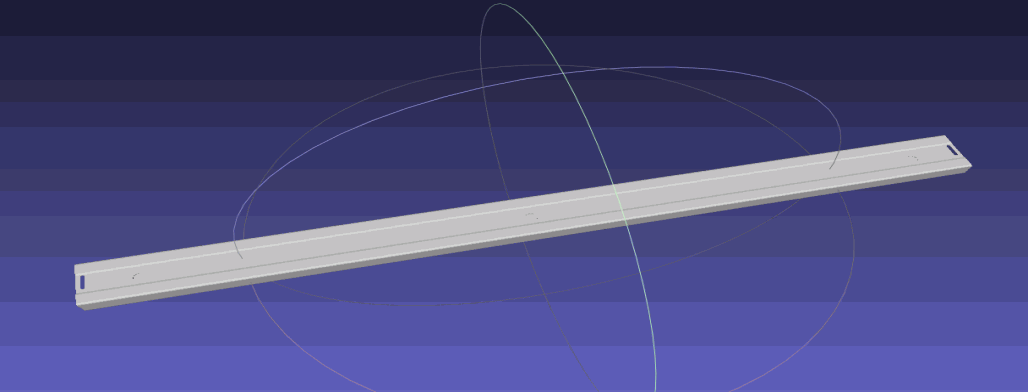}}\\
    \subfigure[PS6D\_7]{
		\includegraphics[width=0.31\linewidth]{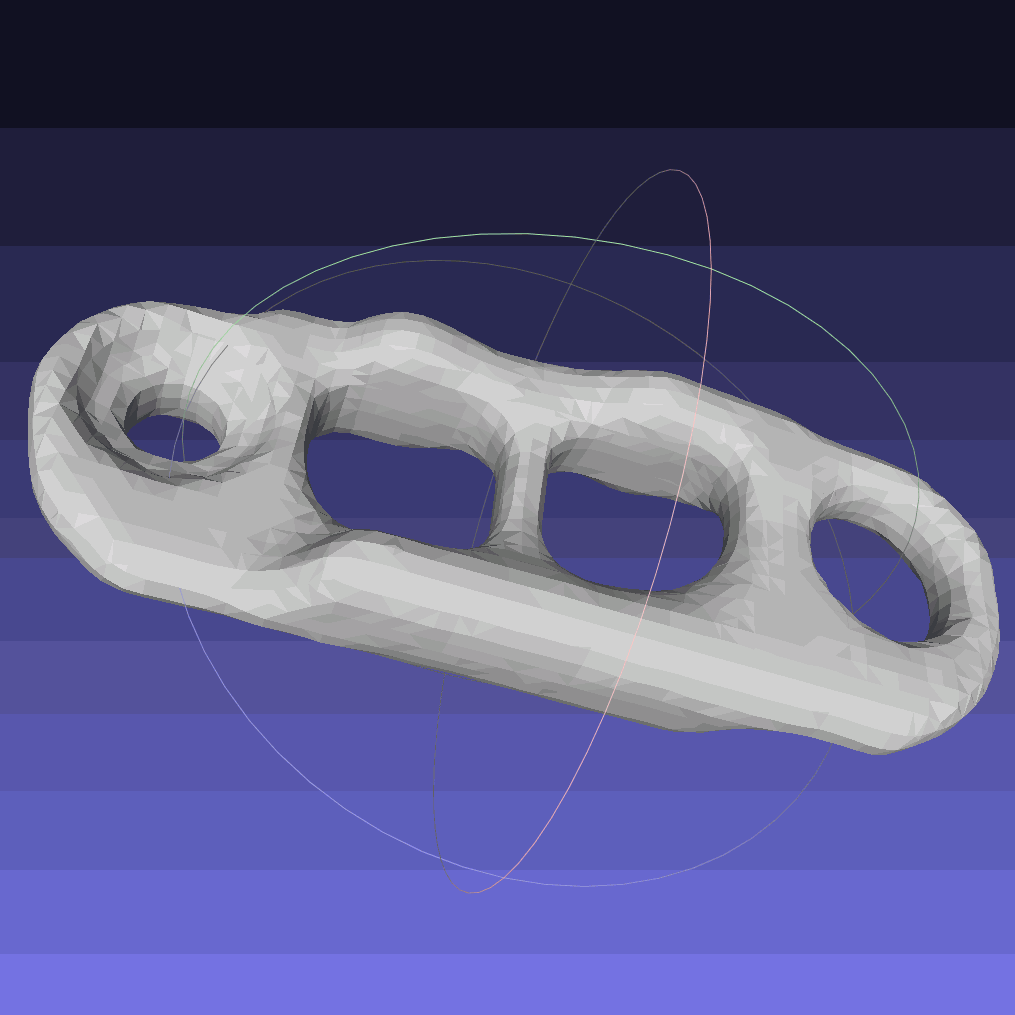}}
	\subfigure[PS6D\_10]{
		\includegraphics[width=0.31\linewidth]{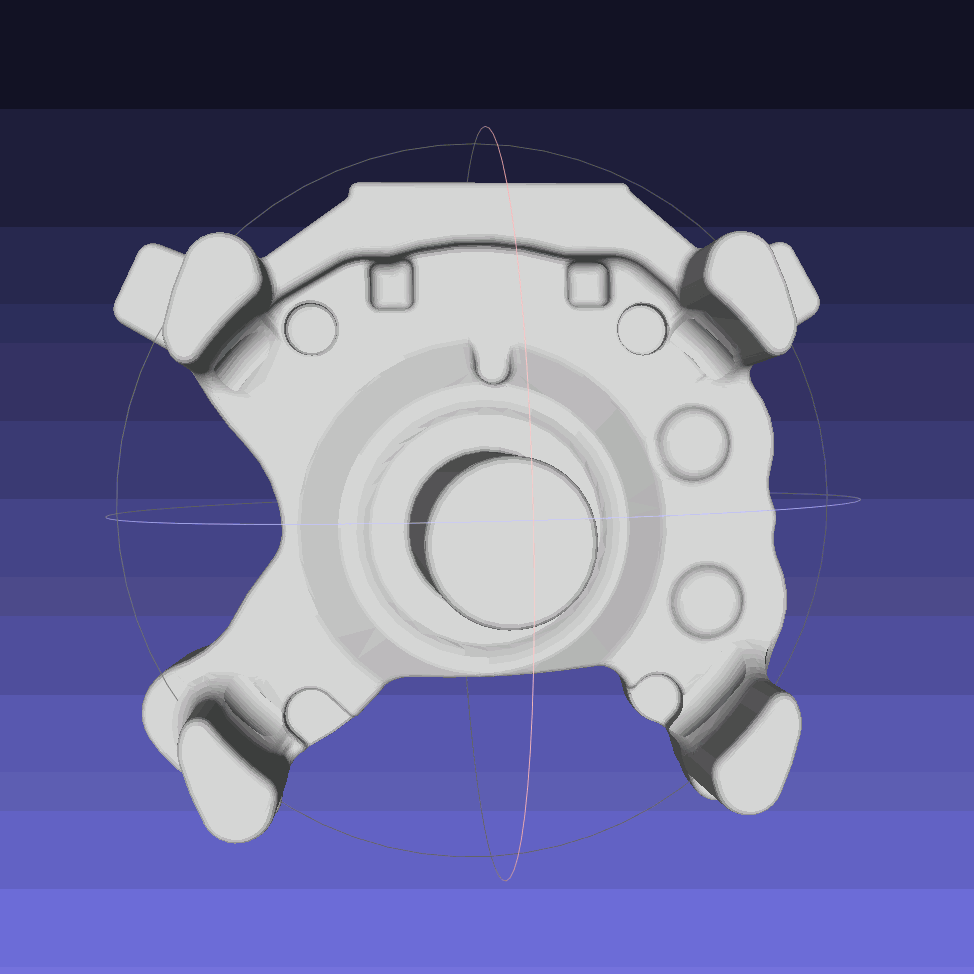}}
    \subfigure[PS6D\_24]{
		\includegraphics[width=0.31\linewidth]{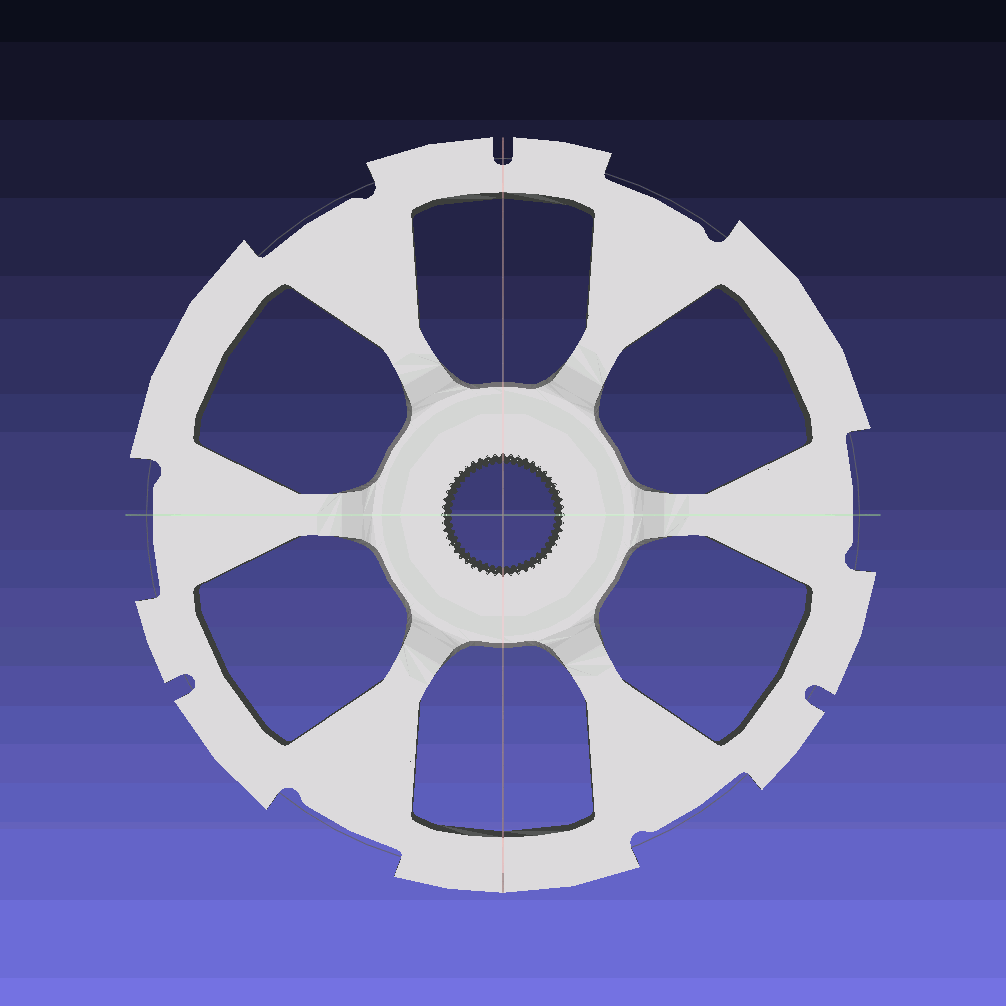}}
	\caption{PS6D Dataset Object Models. Among them, (a), (c), and (d) are objects without symmetry, while (b) and (e) are objects with finite symmetry.}
    \label{PS6D dataset}
\end{figure}

For the selection of objects, we chose Bunny, Tless\_20, Candlestick from the Siléane dataset, and GearShaft and RingScrew from the IPA dataset. Additionally, combining practical considerations in industrial grasping, we utilized the data generation method mentioned above to create PS6D dataset for five common industrial objects as shown in Figs. \ref{PS6D dataset}. PS6D dataset has been published on \href{https://github.com/yangyifanYYF/PS6D\_dataset}{https://github.com/yangyifanYYF/PS6D\_dataset}. 
\subsection{Evaluation Metrics}
When designing evaluation metrics, we aim to achieve the following: 1) The metric should reflect both missing detection and false detection rates, ensuring that the number of detected objects aligns reasonably well with the true number. 2) The metric should reflect point-wise matching accuracy, ensuring that predicted poses closely match the true poses.

The evaluation metric we designed consists of two parts. 

\noindent\textbf{Instance-Level F$_1$-score\quad} F$_{1_{inst}}$ represents the instance-level $F_1$-score, calculated as shown in equation \ref{F1}, where $TP$, $TN$, $FP$, and $FN$ represent the four elements of the confusion matrix. In this context, the $F_1$-score formula is simplified, where $N_{gt}$ represents the number of visible objects in the scene, and $N_{pred}$ represents the predicted number of objects. We consider a predicted pose paired with its nearest true pose as a matched pose. An instance is considered a true positive if the distance is less than $T_e$, where $T_e$ is the tolerance threshold. Considering that heavily occluded objects should not be grasped and should not overly influence the evaluation metrics, we only consider instances with visibility greater than 0.4. Therefore, $N_{gt}$ in the scene can be calculated using equation \ref{Ngt}, where \(n_i\) represents the number of sampled points for the \(i\)-th object, and \(T_v\) is the visibility threshold.
\begin{equation}
\begin{aligned}
    F_{1_{inst}} &= (\cfrac{\frac{TP+FP}{TP}+\frac{TP+FN}{TP}}{2})^{-1}\\
    &= (\cfrac{\frac{N_{pred}}{TP}+\frac{N_{gt}}{TP}}{2})^{-1}
    \label{F1}
\end{aligned}
\end{equation}
\begin{equation}
    N_{gt} = \vert \{ M_i | \frac{n_i}{\max \{ n_i | i \in [1, N_{pred}]\}} > T_v, M_i \in M \} \vert
    \label{Ngt}
\end{equation}
\noindent\textbf{Point-wise Recall\quad}The point-wise recall reflects the matching status of each individual point, and its formula is shown in equation \ref{recall}, representing the difference between the total number of valid points with distances less than the threshold and the total number of points. Here, $D_i$ represents the distance metric, considering symmetry factors, calculated as shown in equation \ref{Di}.
\begin{equation}
    Recall = \frac{\sum\limits_{i=1}^{N_{gt}}|D_i < T_e|}{\sum\limits_{i=1}^{N_{gt}} M_i}
    \label{recall}
\end{equation}
\begin{equation}
    D_i = \min\limits _{s\in S} [(R_i^{gt}s( M_i \cdot v ) + T_i^{gt})- (R_i^{pred}( M_i \cdot v ) +T_i^{pred})]
    \label{Di}
\end{equation}
\subsection{Results}
\begin{table*}[tb]
	\renewcommand\arraystretch{1.2}
 \setlength{\abovecaptionskip}{0pt}    
        \setlength{\belowcaptionskip}{5pt}
	\centering
	\caption{Comparison of different method on 6D pose estimation}
	\resizebox{0.9\textwidth}{!}{
		\begin{tabular}{ccccccccccc}
			\toprule[1.5pt]
			\multirow{2}{*}{Dataset} & \multirow{2}{*}{Object} & \multirow{2}{*}{BBox (mm)} &  \multicolumn{2}{c}{PPF \cite{r1}}&\multicolumn{2}{c}{PPR-Net \cite{r25}} &\multicolumn{2}{c}{PPR-Net++ \cite{r26}}&\multicolumn{2}{c}{PS6D (ours)}\\
         & & &F$_{1_{inst}}$ & Recall & F$_{1_{inst}}$ & Recall & F$_{1_{inst}}$ & Recall & F$_{1_{inst}}$ & Recall \\ \hline
        \multirow{3}{*}{Siléane \cite{5}} & Bunny &[38 41 56] & 32.29 & 30.35 & 98.85 & 95.78 & 99.04 & 98.82 & 99.71 & 99.26 \\ 
        &Tless\_20 &[47 76 83]& 80.83 & 53.76 & 95.74 & 92.27 & 96.28 & 93.65 & 98.94 & 96.19 \\ 
        &Candlestick &[89 89 141]& 49.69 & 45.97 & 95.87 & 97.70 & 96.24 & 98.31 & 100.00 & 100.00 \\ 
        \multirow{2}{*}{IPA \cite{6}} & IPAGearShaft &[99 99 436]& 25.30 & 23.97 & 89.95 & 90.80 & 95.00 & 91.07 & 98.13 & 93.54 \\ 
        & IPARingScrew &[65 99 151]& 60.10 & 50.94 & 96.76 & 89.35 & 96.92 & 92.74 & 99.66 & 99.01 \\ 
        \multirow{5}{*}{PS6D} & PS6D\_6 &[14 18 56]& 42.00 & 27.56
 & 98.43 & 92.37 & 98.52 & 93.70 & 99.98
 & 98.64
 \\ 
        & PS6D\_7 &[38 81 216]& 87.33 & 73.89 & 76.23 & 66.85 & 80.25 & 69.66 & 97.00 & 85.37 \\ 
        & PS6D\_10 &[77 116 125]& 47.45 & 32.19 & 88.85 & 71.93 & 93.22 & 70.24 & 99.70 & 81.85 \\ 
        & PS6D\_23 &[32 92 1304]& 33.31 & 28.80 & 15.22 & 19.05 & 24.47 & 22.10 & 82.92 & 66.1 \\ 
        & PS6D\_24 &[50 240 244]& 57.58 & 47.53 & 91.70 & 69.10 & 95.33 & 70.77 & 99.95 & 99.33 \\ \hline
        \multicolumn{3}{c}{Mean}&51.59&41.50&84.76&78.52&87.53&80.11&97.60&91.93\\
			\bottomrule[1.5pt]
	   \end{tabular}}
	\label{Comparison}
\end{table*}
\begin{figure*}[tb]	
	\centering
  \subfigure[Bunny]{
		\includegraphics[height=0.33\linewidth]{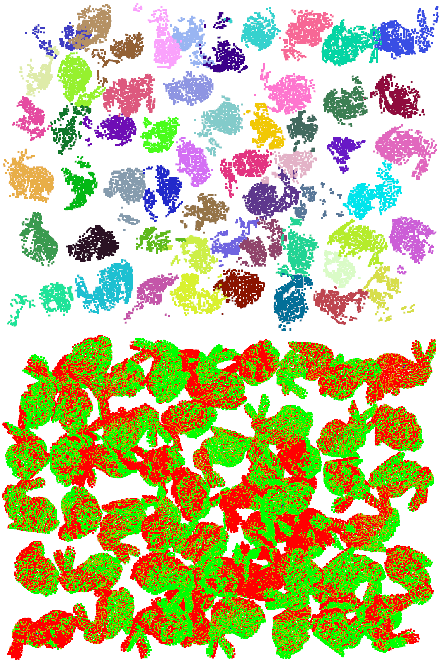}}
  \subfigure[Tless\_20]{
		\includegraphics[height=0.33\linewidth]{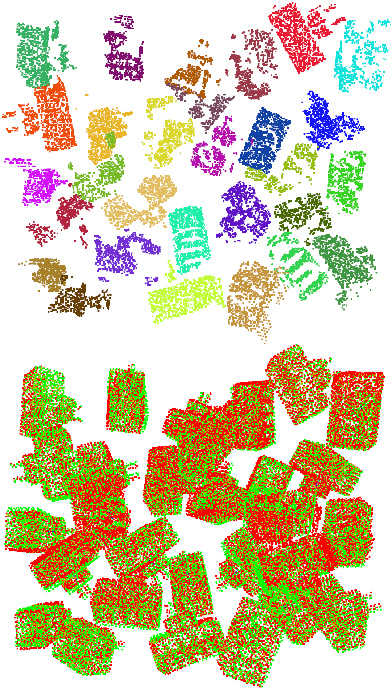}}
  \subfigure[Candlestick]{
		\includegraphics[height=0.33\linewidth]{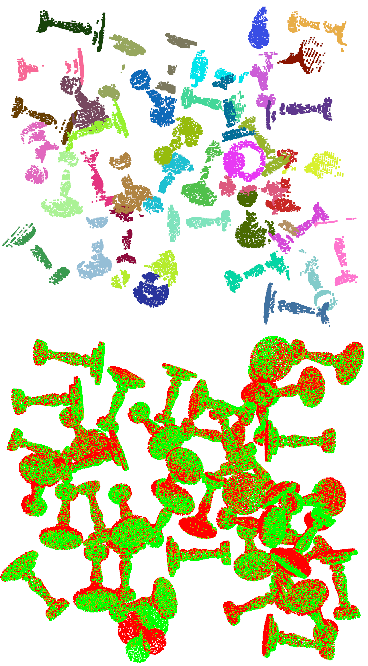}}
  \subfigure[IPAGearShaft]{
		\includegraphics[height=0.33\linewidth]{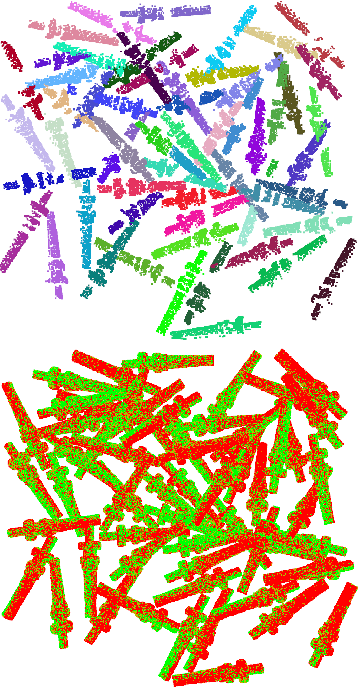}}
  \subfigure[IPARingScrew]{
		\includegraphics[height=0.33\linewidth]{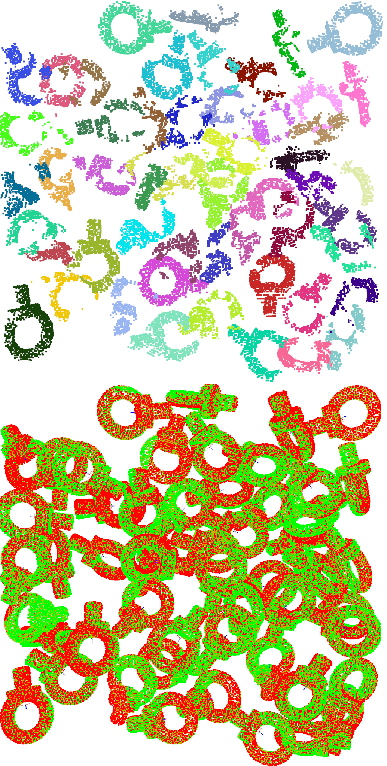}}\\
  \subfigure[PS6D\_6]{
		\includegraphics[height=0.33\linewidth]{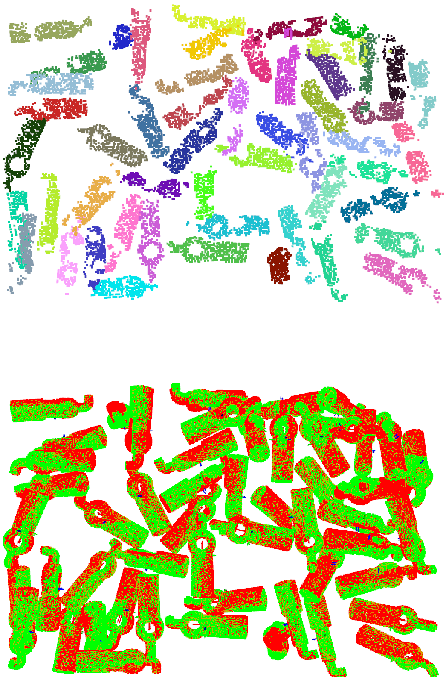}}
  \subfigure[PS6D\_7]{
		\includegraphics[height=0.33\linewidth]{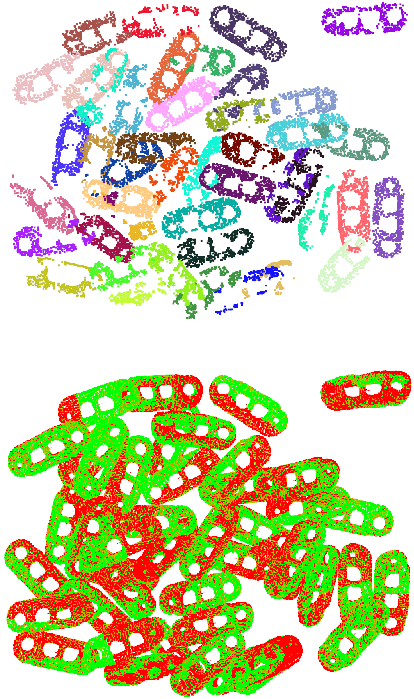}}
  \subfigure[PS6D\_10]{
		\includegraphics[height=0.33\linewidth]{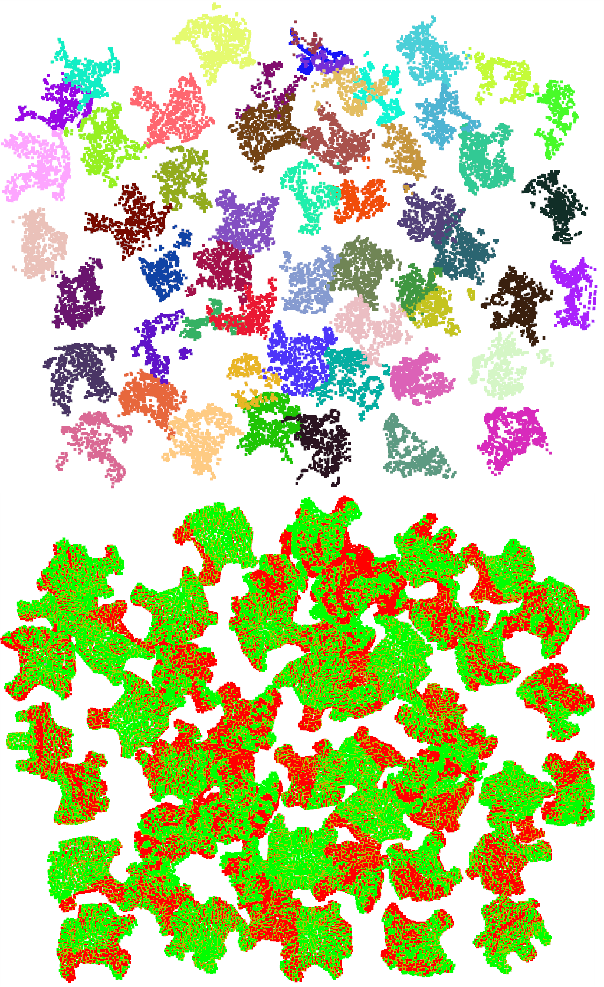}}
  \subfigure[PS6D\_23]{
		\includegraphics[height=0.33\linewidth]{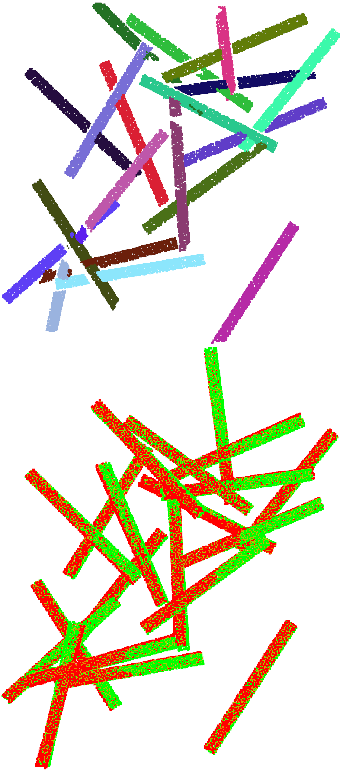}}
  \subfigure[PS6D\_24]{
		\includegraphics[height=0.33\linewidth]{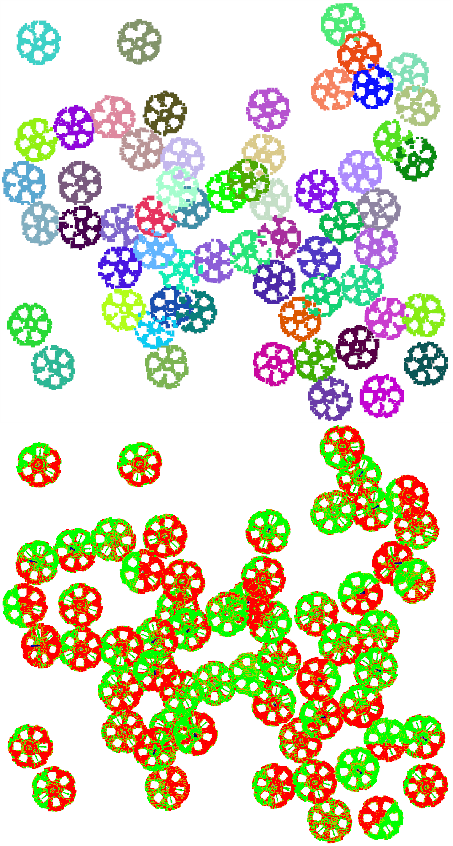}}
	\caption{Simulation visualization results of instance segmentation and pose estimation by PS6D on various objects. In the images above, different colors represent different instances, while in the images below, red point clouds represent predicted poses, and green point clouds represent ground truth poses.}
    \label{visualization}
\end{figure*}

In this part, we will compare PS6D with other state-of-the-art approaches. Due to PS6D relying solely on point cloud information for pose estimation, we need to select some point cloud based methods. PPF \cite{r1} is a commonly used point pair based algorithm in traditional methods. PPR-Net \cite{r25} is a deep neural network algorithm that simultaneously performs instance segmentation and pose estimation, and has achieved good results on the Siléane dataset. PPR-Net++ \cite{r26} is an improved algorithm based on PPR-Net, incorporating bandwidth calculation and confidence learning modules. Table \ref{Comparison} shows the comparison results of PS6D with the three methods on different datasets. Fig. \ref{visualization} illustrates the instance segmentation and pose estimation performance of PS6D on different objects. It should be noted that, when using the PPF algorithm, we provided the number of visible objects in the scene. The tolerance threshold ($T_e$) for all objects, except for PS6D\_23, was set to 5mm. For PS6D\_23, $T_e$ was set to 15mm due to its large size.

The results shows that the PPF algorithm performs mostly poorly in scenarios with occlusion and stacking. For the objects in the Siléane and IPA datasets, PPR-Net and PPR-Net++ have achieved superior performance, but PS6D still has advantages. For the objects in the PS6D dataset, the superiority of PS6D is more pronounced, especially for objects with multiple symmetries (e.g., PS6D\_24) and slender shapes (e.g., PS6D\_23). PS6D shows a significant improvement over other algorithms, with an 11.5\% increase in F$_{1_{inst}}$ and a 14.8\% increase in Recall compared to PPR-Net++.
\begin{figure*}
\begin{minipage}[b]{.5\linewidth}
    \centering
    \subfigure[RGB image]{
		\includegraphics[height=0.3\linewidth]{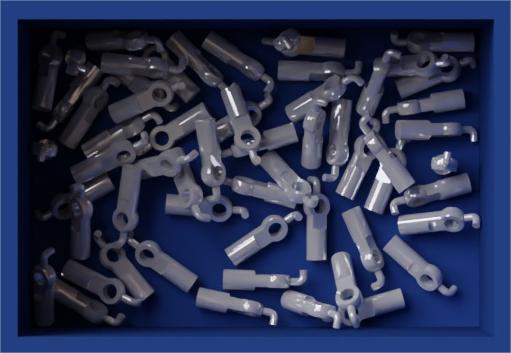}}
    \subfigure[Point cloud]{
		\includegraphics[height=0.3\linewidth]{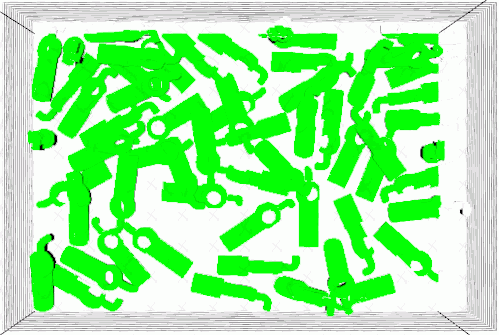}}\\
    \subfigure[Pose estimation]{
		\includegraphics[height=0.3\linewidth]{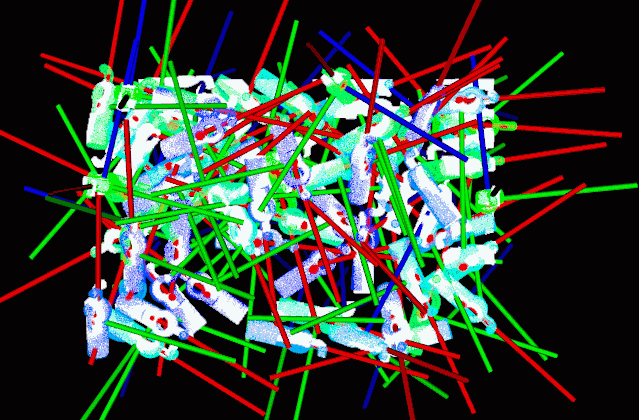}}
  \subfigure[Grasping points]{
		\includegraphics[height=0.3\linewidth]{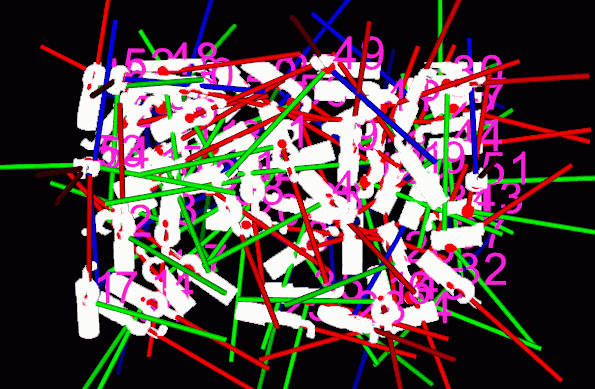}}
\end{minipage} 
\medskip
\begin{minipage}[b]{.5\linewidth}
    \centering
    \subfigure[Picking scene]{
		\includegraphics[height=0.71\linewidth]{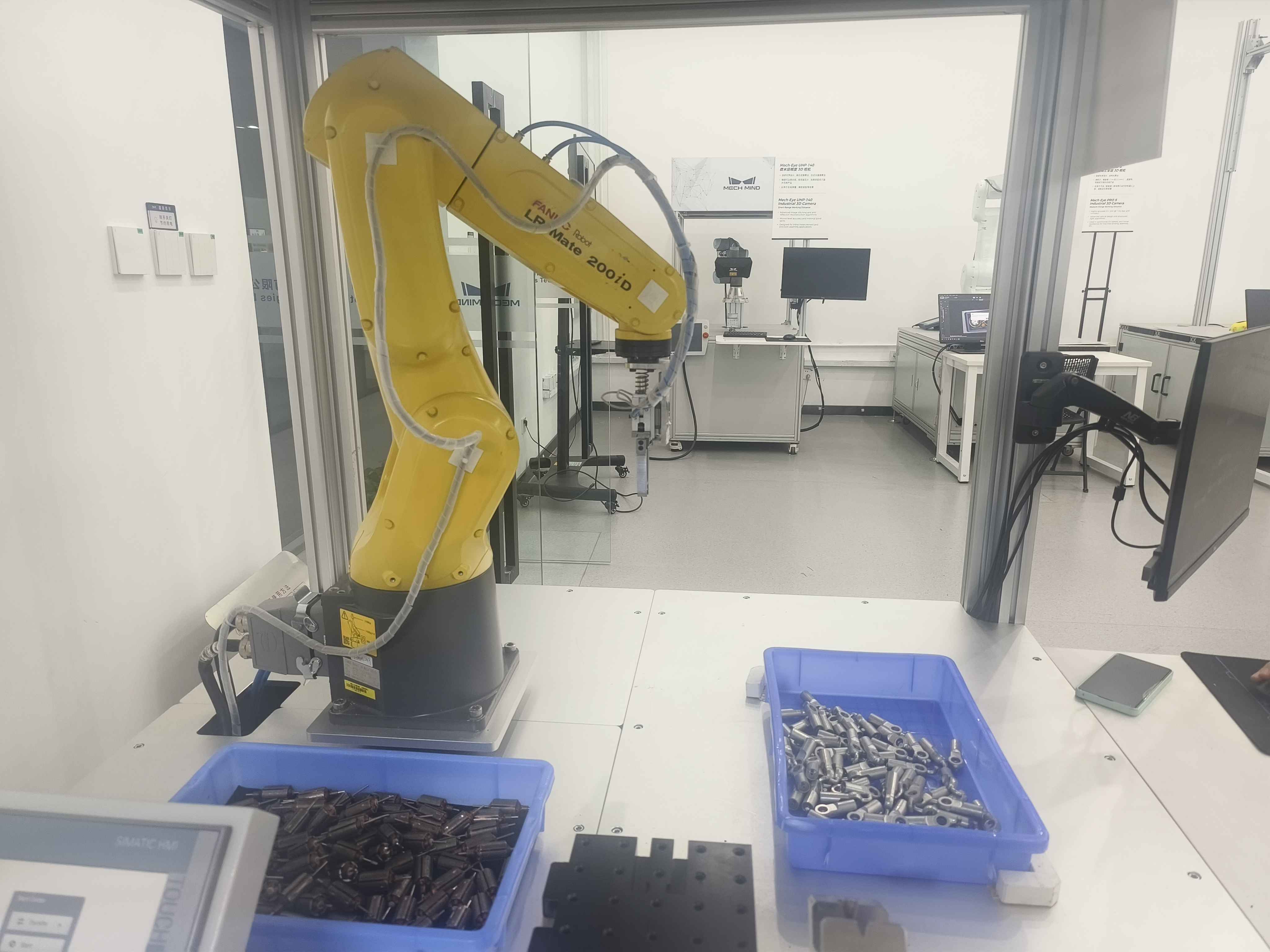}}
\end{minipage}
\caption{The pipeline of our method. 
First, use a camera to capture the scene point cloud and extract the point cloud within the Region of Interest (ROI). Then, use PS6D for instance segmentation and pose estimation, followed by connecting to fine matching for ICP iteration to obtain the grasp point pose. Finally, employ the robotic arm to grasp and place the object. }
    \label{in}
\end{figure*}
\subsection{Ablation Study}
We replace the network backbone with Pointnet2MSG, remove the normalization module, and use single-stage clustering to conduct three ablation experiments. We specifically evaluate the results for three representative objects: PS6D\_23 (slender), PS6D\_24 (multi-symmetrical), and Candlestick (infinite symmetrical), with corresponding metrics provided in Table \ref{ablation study}.
\begin{table}[tb]
	\renewcommand\arraystretch{1.2}
 \setlength{\abovecaptionskip}{0pt}    
        \setlength{\belowcaptionskip}{5pt}
	\centering
	\caption{Results of ablation study}
	\resizebox{0.49\textwidth}{!}{
		\begin{tabular}{ccccccc}
			\toprule[1.5pt]
			~ &\multicolumn{2}{c}{Pointnet2MSG}& \multicolumn{2}{c}{Without Normalization}&\multicolumn{2}{c} {Single Stage Clustering} \\
        ~ & F$_{1_{inst}}$ & Recall & F$_{1_{inst}}$ & Recall & F$_{1_{inst}}$ & Recall \\ \hline
        PS6D\_23 & 64.15 & 45.45 & 2.76 & 4.20 & 77.47 & 59.80 \\ 
        PS6D\_24 & 99.94 & 97.32 & 96.63 & 96.20 & 98.80 & 30.47 \\ 
        Candlestick & 100.00 & 99.97 & 99.95 & 99.97 & 98.71 & 100.00 \\
			\bottomrule[1.5pt]
	   \end{tabular}}
	\label{ablation study}
\end{table}

The results of the ablation experiments indicate that, the transformer structure in the backbone can improve the matching performance to some extent, with more significant improvements for objects with initially lower scores; normalization operations enable the network to adapt to objects of different sizes simultaneously, and the network without normalization performs well only on small-sized objects; two-stage clustering significantly improves the performance for objects with multiple symmetries and slender shapes, while its impact on objects with infinite symmetry is less pronounced.

  

\subsection{Real World Experiment on Robot Bin-picking}
In practical grasping experiments, our objective is to train the PS6D algorithm on a simulated dataset with known CAD models of the objects, utilize the obtained weights to perform inference on point clouds from real scenes, and achieve instance segmentation and pose estimation for grasping pose determination. The pipeline is illustrated in Fig. \ref{in}. In these specific experiments, we use actual PS6D\_6 objects rather than those obtained through 3D printing. These objects have consistent weight and material with those used in industrial settings. They are haphazardly stacked in boxes, simulating the real stacking environment of industrial parts. The Mech-Eye ProS industrial 3D camera captures images and collects point clouds. The scene point cloud within the region of interest (ROI) is input into PS6D without prior 2D segmentation preprocessing. The point cloud undergoes feature extraction, centroid regression, and rotation prediction. Ultimately, the poses of each workpiece are obtained through a two-stage clustering algorithm. The poses of each grasping point are determined using pre-defined grasp points. The precise poses of each grasping point are refined through iterative closest point (ICP) refinement. The Mech-Viz robotic guidance software guides a Fanuc-lr-mate-200id robotic arm to accurately grasp and place the PS6D\_6 object into an adjacent bin. In scenarios involving scattered and stacked objects, our grasping success rate reaches 91.7\%. For the video of robot bin-picking experiments with PS6D, please refer to \href{https://youtu.be/xUvESexyv9c}{https://youtu.be/xUvESexyv9c}.



\section{Conclusion}
This paper introduces a method for instance segmentation and 6D pose estimation based solely on point cloud data. We propose a novel and lightweight network architecture that conducts feature extraction, centroid regression, and rotation prediction on normalized point clouds. Utilizing a two-stage clustering approach for instance segmentation and pose voting, our method provides precise pose predictions with millimeter-level accuracy. PS6D outperforms the state-of-art approaches in 6D pose estimation by 11.5\% in F$_{1_{inst}}$ and 14.8\% Recall, attaining a 91.7\% success rate in real-world bin-picking tests. The proposed approach demonstrates accurate pose estimation in challenging industrial grasping scenarios characterized by textureless, reflective, and dark surfaces. This capability contributes to the seamless execution of subsequent grasping tasks in industrial applications. The pose estimation results obtained through our method are applicable to industrial grasping. 
\section*{ACKNOWLEDGMENT}
We express our sincere gratitude to the support provided by Mech-Mind, including access to abundant data resources, advanced machinery equipment, and comprehensive software solutions. This work is supported by National Natural Science Foundation of China under Grant 62173189.



\begin{thebibliography}{99}
\bibitem{r1} B. Drost, M. Ulrich, N. Navab, and S. Ilic, “Model globally, match locally: Efficient and robust 3d object recognition,” in Computer Vision and Pattern Recognition (CVPR), 2010 IEEE Conference on. Ieee, 2010, pp. 998–1005.
\bibitem{r2}S. Hinterstoisser, V. Lepetit, S. Ilic, S. Holzer, G. Bradski, K. Konolige, et al., "Model based training detection and pose estimation of texture-less 3d objects in heavily cluttered scenes", Asian conference on computer vision, pp. 548-562, 2012.
\bibitem{r6}W. Kehl, F. Manhardt, F. Tombari, S. Ilic and N. Navab, "SSD-6D: Making RGB-Based 3D Detection and 6D Pose Estimation Great Again," 2017 IEEE International Conference on Computer Vision (ICCV), Venice, Italy, 2017, pp. 1530-1538.
\bibitem{r7}Y. Xiang, T. Schmidt, V. Narayanan and D. Fox, "Posecnn: A convolutional neural network for 6d object pose estimation in cluttered scenes", Robotics: Science and Systems (RSS), 2018.
\bibitem{r8}B. Tekin, S. N. Sinha and P. Fua, "Real-Time Seamless Single Shot 6D Object Pose Prediction," 2018 IEEE/CVF Conference on Computer Vision and Pattern Recognition, Salt Lake City, UT, USA, 2018, pp. 292-301.
\bibitem{r9}S. Peng, Y. Liu, Q. Huang, X. Zhou and H. Bao, "PVNet: Pixel-Wise Voting Network for 6DoF Pose Estimation," 2019 IEEE/CVF Conference on Computer Vision and Pattern Recognition (CVPR), Long Beach, CA, USA, 2019, pp. 4556-4565.
\bibitem{r10}S. H. Bengtson, H. Åström, T. B. Moeslund, E. A. Topp and V. Krueger, "Pose Estimation from RGB Images of Highly Symmetric Objects using a Novel Multi-Pose Loss and Differential Rendering," 2021 IEEE/RSJ International Conference on Intelligent Robots and Systems (IROS), Prague, Czech Republic, 2021, pp. 4618-4624.
\bibitem{r11}T. Hodaň, D. Baráth and J. Matas, "EPOS: Estimating 6D Pose of Objects With Symmetries," 2020 IEEE/CVF Conference on Computer Vision and Pattern Recognition (CVPR), Seattle, WA, USA, 2020, pp. 11700-11709.
\bibitem{r12}D. Xu, D. Anguelov and A. Jain, "PointFusion: Deep Sensor Fusion for 3D Bounding Box Estimation," 2018 IEEE/CVF Conference on Computer Vision and Pattern Recognition, Salt Lake City, UT, USA, 2018, pp. 244-253.
\bibitem{pointnet}C. R. Qi, H. Su, K. Mo and L. J. Guibas, "Pointnet: Deep learning on point sets for 3d classification and segmentation", Proc. Computer Vision and Pattern Recognition (CVPR), vol. 1, no. 2, pp. 4, 2017.
\bibitem{r13}C. Wang et al., "DenseFusion: 6D Object Pose Estimation by Iterative Dense Fusion," 2019 IEEE/CVF Conference on Computer Vision and Pattern Recognition (CVPR), Long Beach, CA, USA, 2019, pp. 3338-3347.
\bibitem{r14}Y. He, W. Sun, H. Huang, J. Liu, H. Fan and J. Sun, "PVN3D: A Deep Point-Wise 3D Keypoints Voting Network for 6DoF Pose Estimation," 2020 IEEE/CVF Conference on Computer Vision and Pattern Recognition (CVPR), Seattle, WA, USA, 2020, pp. 11629-11638.
\bibitem{r15}Y. He, H. Huang, H. Fan, Q. Chen and J. Sun, "FFB6D: A Full Flow Bidirectional Fusion Network for 6D Pose Estimation," 2021 IEEE/CVF Conference on Computer Vision and Pattern Recognition (CVPR), Nashville, TN, USA, 2021, pp. 3002-3012.
\bibitem{r16}N. Mo, W. Gan, N. Yokoya and S. Chen, "ES6D: A Computation Efficient and Symmetry-Aware 6D Pose Regression Framework," 2022 IEEE/CVF Conference on Computer Vision and Pattern Recognition (CVPR), New Orleans, LA, USA, 2022, pp. 6708-6717.
\bibitem{r17}Y. Di et al., "GPV-Pose: Category-level Object Pose Estimation via Geometry-guided Point-wise Voting," 2022 IEEE/CVF Conference on Computer Vision and Pattern Recognition (CVPR), New Orleans, LA, USA, 2022, pp. 6771-6781.
\bibitem{r18}J. Lin, Z. Wei, Z. Li, S. Xu, K. Jia and Y. Li, "DualPoseNet: Category-level 6D Object Pose and Size Estimation Using Dual Pose Network with Refined Learning of Pose Consistency," 2021 IEEE/CVF International Conference on Computer Vision (ICCV), Montreal, QC, Canada, 2021, pp. 3540-3549.
\bibitem{r19}F. Duffhauss, S. Koch, H. Ziesche, N. A. Vien and G. Neumann, "SyMFM6D: Symmetry-Aware Multi-Directional Fusion for Multi-View 6D Object Pose Estimation," in IEEE Robotics and Automation Letters, vol. 8, no. 9, pp. 5315-5322, Sept. 2023.
\bibitem{r20}D. -C. Hoang, J. A. Stork and T. Stoyanov, "Voting and Attention-Based Pose Relation Learning for Object Pose Estimation From 3D Point Clouds," in IEEE Robotics and Automation Letters, vol. 7, no. 4, pp. 8980-8987, Oct. 2022.
\bibitem{r21}D. Cai, J. Heikkiä and E. Rahtu, "OVE6D: Object Viewpoint Encoding for Depth-based 6D Object Pose Estimation," 2022 IEEE/CVF Conference on Computer Vision and Pattern Recognition (CVPR), New Orleans, LA, USA, 2022, pp. 6793-6803.
\bibitem{r22}K. Kleeberger and M. F. Huber, "Single Shot 6D Object Pose Estimation," 2020 IEEE International Conference on Robotics and Automation (ICRA), Paris, France, 2020, pp. 6239-6245.
\bibitem{r23}L. Zeng, W. J. Lv, X. Y. Zhang and Y. J. Liu, "ParametricNet: 6DoF Pose Estimation Network for Parametric Shapes in Stacked Scenarios," 2021 IEEE International Conference on Robotics and Automation (ICRA), Xi'an, China, 2021, pp. 772-778.
\bibitem{r24}Y. Li, T. Kong, R. Chu, Y. Li, P. Wang and L. Li, "Simultaneous Semantic and Collision Learning for 6-DoF Grasp Pose Estimation," 2021 IEEE/RSJ International Conference on Intelligent Robots and Systems (IROS), Prague, Czech Republic, 2021, pp. 3571-3578.
\bibitem{r25}Z. Dong et al., "PPR-Net:Point-wise Pose Regression Network for Instance Segmentation and 6D Pose Estimation in Bin-picking Scenarios," 2019 IEEE/RSJ International Conference on Intelligent Robots and Systems (IROS), Macau, China, 2019, pp. 1773-1780.
\bibitem{r26}L. Zeng, W. J. Lv, Z. K. Dong and Y. J. Liu, "PPR-Net++: Accurate 6-D Pose Estimation in Stacked Scenarios," in IEEE Transactions on Automation Science and Engineering, vol. 19, no. 4, pp. 3139-3151.
\bibitem{1}C. R. Qi, L. Yi, H. Su, and L. J. Guibas, "Pointnet++: Deep hierarchical feature learning on point sets in a metric space," in Advances in Neural Information Processing Systems, 2017, pp. 5099–5108.
\bibitem{2} H. Zhao, L. Jiang, J. Jia, P. Torr and V. Koltun, "Point Transformer," 2021 IEEE/CVF International Conference on Computer Vision (ICCV), Montreal, QC, Canada, 2021, pp. 16239-16248.
\bibitem{3}S. Hinterstoisser, S. Holzer, C. Cagniart, S. Ilic, K. Konolige, N. Navab, et al., "Multimodal templates for real-time detection of texture-less objects in heavily cluttered scenes", Proceedings of the IEEE International Conference on Computer Vision (ICCV), pp. 858-865, 2011.
\bibitem{4}T. Hodan, P. Haluza, Š. Obdržálek, J. Matas, M. Lourakis and X. Zabulis, "T-LESS: An RGB-D Dataset for 6D Pose Estimation of Texture-Less Objects," 2017 IEEE Winter Conference on Applications of Computer Vision (WACV), Santa Rosa, CA, USA, 2017, pp. 880-888.
\bibitem{5}R. Brégier, F. Devernay, L. Leyrit and J. L. Crowley, "Symmetry aware evaluation of 3d object detection and pose estimation in scenes of many parts in bulk", 2017 IEEE International Conference on Computer Vision Workshop (ICCVW), pp. 2209-2218, 2017.
\bibitem{6}K. Kleeberger, C. Landgraf and M. F. Huber, "Large-scale 6D object pose estimation dataset for industrial bin-picking", Proc. IEEE/RSJ Int. Conf. Intell. Robots Syst. (IROS), pp. 2573-2578, Nov. 2019.

\end{thebibliography}
\end{document}